\documentclass{article}

\usepackage[preprint,nonatbib]{neurips_2023}

\usepackage[colorlinks,citecolor=gray,linkcolor=red]{hyperref} 

\usepackage[utf8]{inputenc} 
\usepackage[T1]{fontenc}    
\usepackage{hyperref}       
\usepackage{url}            
\usepackage{booktabs}       
\usepackage{amsfonts}       
\usepackage{nicefrac}       
\usepackage{microtype}      
\usepackage{xcolor}         
\usepackage{multirow}

\usepackage{graphicx}
\usepackage{multirow}
\usepackage{subcaption}
\usepackage{dblfloatfix}
\usepackage{booktabs}
\usepackage{colortbl}
\usepackage{enumitem}

\usepackage{times}
\usepackage{epsfig}
\usepackage{graphicx}
\usepackage{amsmath}
\usepackage{amssymb}
\usepackage{xspace}
\usepackage{xcolor}
\usepackage{cleveref}

\definecolor{mygray}{gray}{.9}
\definecolor{mygreen}{RGB}{34,139,34}

\newcommand{\cls}[1]{{\color{blue}{\@#1}}}
\newcommand{\dete}[1]{{\color{purple}{\@#1}}}
\newcommand{\seg}[1]{{\color{mygreen}{\@#1}}}

\newcommand{\model}{MTHL\xspace}
\newcommand{\eg}{\emph{e.g.}}

\title{An Efficient General-Purpose Modular Vision Model via Multi-Task Heterogeneous Training}

\author{
Zitian Chen\textsuperscript{1}, Mingyu Ding\textsuperscript{2}, Yikang Shen\textsuperscript{3},  Wei Zhan\textsuperscript{2}, \\
\textbf{ Masayoshi Tomizuka\textsuperscript{2}, Erik Learned-Miller\textsuperscript{1}, Chuang Gan\textsuperscript{1,3} } \\
        {
            \small \textsuperscript{1} University of Massachusetts Amherst, \textsuperscript{2} University of California Berkeley, \textsuperscript{3} MIT-IBM Watson AI Lab
        }
        }
        
\begin{document}

\maketitle

\begin{abstract}

We present a model that can perform multiple vision tasks and can be adapted to other downstream tasks efficiently. 
Despite considerable progress in multi-task learning, most efforts focus on learning from \textit{multi-label data}: a single image set with multiple task labels.
Such multi-label data sets are rare, small, and expensive.  We say \textit{heterogeneous} to refer to image sets with different task labels, or to combinations of single-task datasets. Few have explored training on such heterogeneous datasets. General-purpose vision models are still dominated by single-task pretraining, and it remains unclear how to scale up multi-task models by leveraging mainstream vision datasets designed for different purposes. The challenges lie in managing large intrinsic differences among vision tasks, including data distribution, architectures, task-specific modules, dataset scales, and sampling strategies. To address these challenges, we propose to modify and scale up mixture-of-experts (MoE) vision transformers, so that they can simultaneously learn classification, detection, and segmentation on diverse mainstream vision datasets including ImageNet, COCO, and ADE20K. Our approach achieves comparable results to single-task state-of-the-art models and demonstrates strong generalization on downstream tasks. Due to its emergent modularity, this general-purpose model decomposes into high-performing components, efficiently adapting to downstream tasks. We can fine-tune it with fewer training parameters, fewer model parameters, and less computation.
Additionally, its modularity allows for easy expansion in continual-learning-without-forgetting scenarios. Finally, these functions can be controlled and combined to meet various demands of downstream tasks.

\end{abstract}

\section{Introduction}

\begin{figure}[t]
  \centering
  \includegraphics[width=14cm]{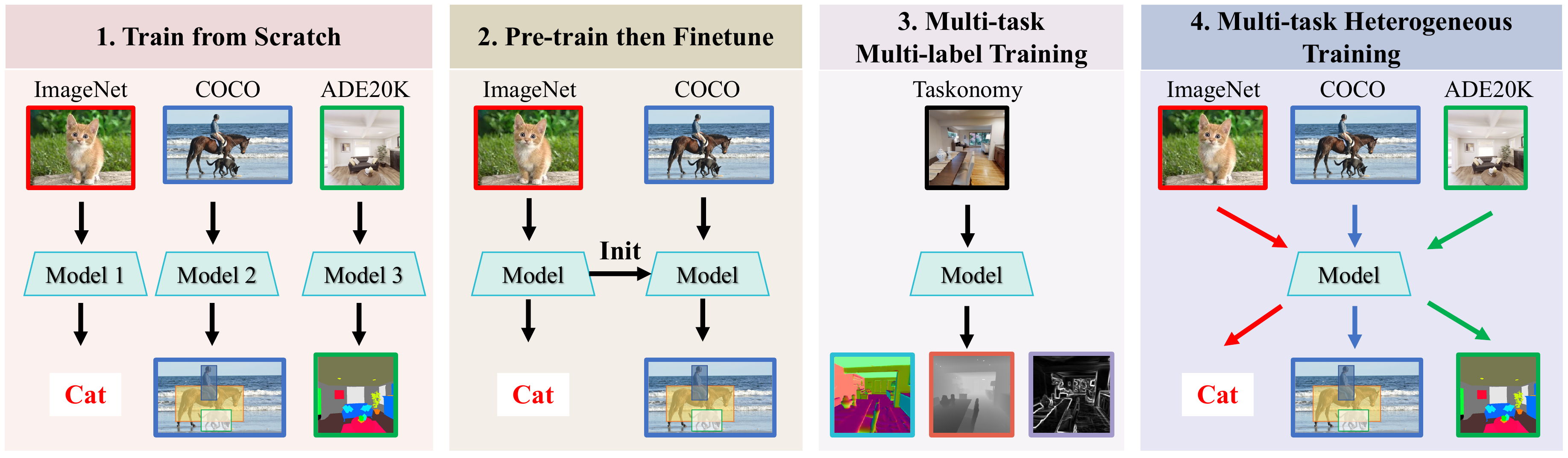}
  \caption{\textbf{Different ways of training.} (1) Train from Scratch: Train a model for a  single task from scratch. (2) Pre-train then Finetune: Pre-train a model on one dataset and later fine-tune the model on other datasets. (3) Multi-task Multi-label Training: Train a model that can produce multiple types of outputs simultaneously. The dataset is expected to have multiple annotations for different tasks on each training image. (4) Multi-task Heterogeneous Training (MTHT): Train a model that can produce different types of outputs corresponding to each task. The model can make use of training data designed for any single task. It can use these in combination to achieve multi-task training.
  }
  \vspace{-18pt}
  \label{fig:training}
\end{figure}

Comprehensive visual understanding demands a general-purpose model capable of performing diverse vision tasks. With a similar goal, multitask learning (MTL), which enables the simultaneous training of models on multiple tasks and allows them to leverage shared information, has been explored extensively. Most MTL efforts~\cite{chen2022modsquad,xumtformer,maninis2019attentive} 
have been made by learning from multi-label datasets, where each input has multiple different types of annotations. However, such data sets with multiple annotations are often impractical to obtain. And the mainstream classification, detection, and segmentation datasets (ImageNet~\cite{deng2009imagenet}, COCO~\cite{COCO}, and ADE20K~\cite{zhou2017scene}) have no overlapping images. 
Hence the current paradigm for general-purpose vision models is still dominated by single-task pretraining (\eg, image classification~\cite{liu2021swin}, self-distillation~\cite{caron2021emerging}, or multi-modal contrastive learning~\cite{yuan2021florence}) and then fine-tuning on downstream tasks. 
A detailed demonstration of different schemes of pre-training is shown in Fig.~\ref{fig:training}. 



The previous work Mod-Squad~\cite{chen2022modsquad} proposes to use a mixture-of-experts (MoE) and a mutual information loss to address task conflict in MTL. However, it oversimplifies some task-specific network designs and the success of this model heavily relies on multi-label datasets, which are difficult to obtain and scale up. Therefore, it remains unclear:
1) How to scale up this MTL model for multi-task heterogeneous training on conventional computer vision datasets;
2) Whether this model can be utilized as a general-purpose vision backbone that can be easily adapted to many downstream tasks;
3) Whether we can leverage the success of single-task methods instead of removing complicated modules and simplifying the task-specific sub-network.


Another issue is that previous large-scale vision models~\cite{chen2022modsquad,liu2021swin,yuan2021florence,caron2021emerging} do not consider fast adaptation on downstream tasks. One common limitation of large-scale models is that adapting to downstream tasks requires updating all parameters, which can be prohibitively expensive in terms of time and computational resources. For example, large models like GPT-3~\cite{brown2020language} with 175B parameters, can take months to train, making adapting the whole model for a tiny new task impractical. Therefore, efficient adaptation is an important practical feature for successful model deployment.

To address these problems, we build a large-scale multi-task heterogeneous training framework based on a modular vision transformer that can simultaneously do three fundamental vision tasks: classification, detection, and segmentation. We refer to this framework as {\bf Multi-task Heterogeneous Learner (MTHL)}. 
Benefiting from a more diverse training set designed for multiple purposes, the framework generalizes better and is semantically rich enough for rapid downstream adaptation, which is often hard to obtain from a single (homogeneous) pre-training task/dataset. 

\begin{figure}[t]
  \centering
  \includegraphics[width=14cm]{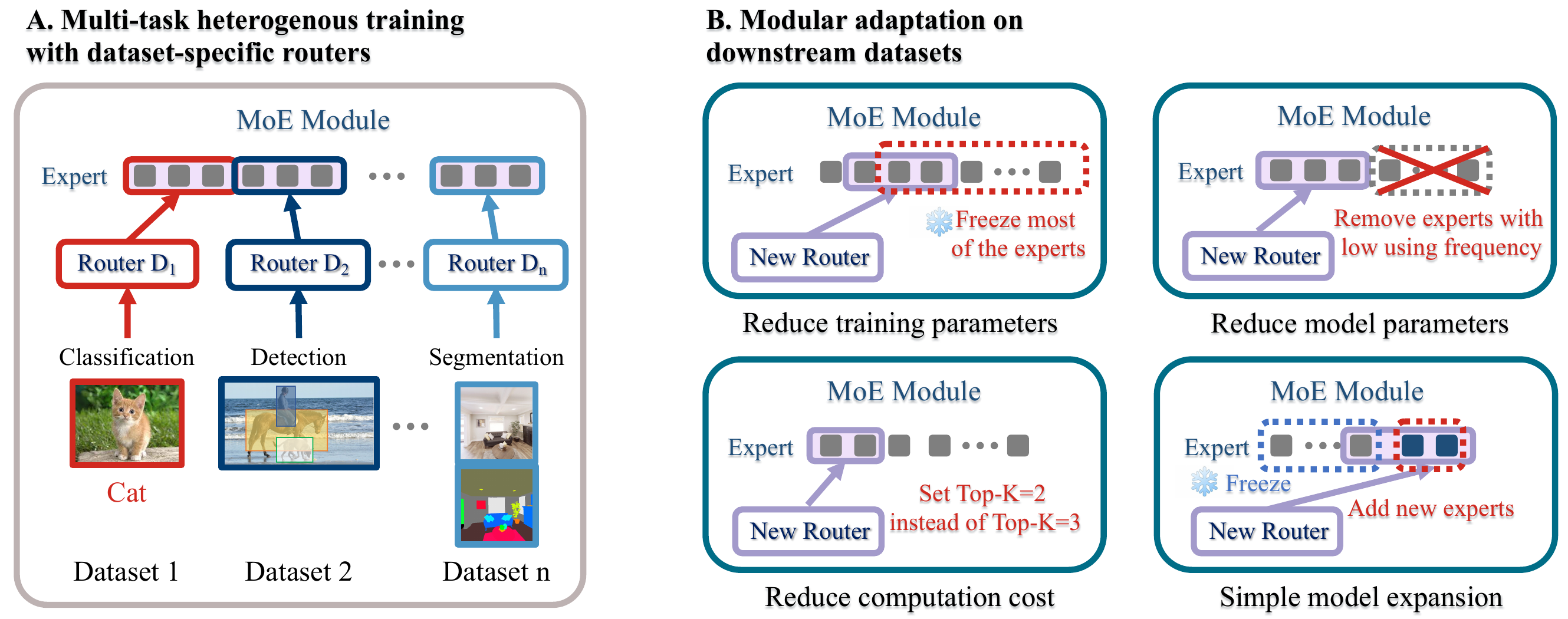}
  \caption{\textbf{Efficient modular adaption.} 
  Strong modularity facilitates efficient adaptation to new datasets:
  1) Reduce training parameters by only learning new routers and a few optional experts while freezing other parameters. 2) Reduce model parameters via learning new routers and removing rarely selected experts. 3) Reduce computation cost via learning new routers with a smaller Top-K that makes fewer experts chosen in one forward pass. 4) Simple model expansion via adding and learning a few new experts per MoE module while freezing old experts. Above ways of adaptation can be combined to suit specific needs. 
  }
  \vspace{-10pt}
  \label{fig:adapt}
\end{figure}

We also address efficient adaptation by leveraging the strong modularity in our model. As shown in Fig.~\ref{fig:adapt}, \model can adapt efficiently in several aspects including reducing training parameters, model parameters, and computational cost. The mixture-of-experts module enables the model to select the most semantically meaningful part for faster transferring to downstream tasks by simply learning new routers. Further, the model can easily expand by adding experts to address continual learning.

Our main contributions can be summarized as follows:
\begin{itemize}[align=right,itemindent=0em,labelsep=5pt,labelwidth=1em,leftmargin=20pt,itemsep=0em] 
\vspace{-6pt}
\item \textbf{Large-scale multi-task heterogenous training.} We explore heterogeneous training on three fundamental computer vision tasks: classification, detection, and segmentation with mainstream vision datasets. We demonstrate that one model to perform three tasks can be on par with single-task state-of-the-art.

\item \textbf{Strong generalization on downstream datasets.}
Heterogeneous training provides the advantage of diverse perception and the ability to handle a wider range of scenarios, leading to better generalization to downstream datasets.

\item \textbf{Modular adaptation with efficiency.}  The emergence of modularity allows for flexible control of architecture and provides several straightforward and efficient ways to adapt the architecture.


\item \textbf{Continual learning without forgetting.} The model can effortlessly leverage existing experts to adapt to new tasks by learning new routers. Additionally, it can incorporate new experts without disrupting the current architecture, thus avoiding catastrophic forgetting.



\end{itemize}


\section{Related Work}

\noindent \textbf{Multi-task Learning.} Multi-task learning~\cite{kendall2018multi} jointly learns multiple related tasks with a single model. Recently, transformer-based MTL architectures~\cite{xumtformer} have gained popularity. Some works~\cite{jaegle2021perceiver, lu2023unifiedio} attempt to unify the input and output space for different tasks. 
Some works~\cite{chen2022modsquad,xumtformer,maninis2019attentive} remove complicated task-specific module for simplicity and conducts multi-task learning on a multi-label dataset.
However, these works either rely on a single multi-label dataset or lose some perceptual feature when removing the task-specific modules and unifying the input/output~\cite{jaegle2021perceiver, lu2023unifiedio}.
While Ubernet\cite{kokkinos2017ubernet} adapts a CNN-based framework that can learn from multiple datasets, it struggles with task conflicts and tends to have lower performance when learning from more than one dataset, which makes it hard to generalize to downstream applications.
In contrast, \model can learn from diverse datasets and still achieve comparable performance to state-of-the-art single-task methods. Additionally, it can leverage the success of single-task methods by adopting similar designs, including unique task-specific modules (e.g., anchor generator), data pre-processing, and complicated engineering techniques (e.g., non-maximum suppression). These designs are non-trivial but necessary to achieve the best model.



\noindent \textbf{Mixture of Experts (MoE).} 
Jacobs et al.~\cite{jacobs1991adaptive} introduced the MoE as a method to merge sub-models and perform conditional computation. Recently, this technique has been commonly used to decrease computation costs while maintaining a large model capacity~\cite{shazeer2017}. Some studies~\cite{lepikhin2021gshard, JMLR:v23:21-0998,riquelme2021scaling, mustafa2022multimodal} have leveraged MoE to train massive models with trillions of parameters at a relatively low computation cost. In contrast, we utilize this technique primarily to manage the sub-models and conduct the modular adaptation on downstream tasks.

\noindent \textbf{Parameter-efficient transfer learning.} 
The Adapter technique was proposed as a standalone layer that can be integrated into an existing neural network for efficient transfer. LoRA~\cite{hu2021lora} utilizes a bottleneck structure to enforce a low-rank constraint on the weight updates. Other approaches integrate CLIP-based adapters~\cite{gao2021clip, sung2022vladapter, zhang2021tip}, upsampling and downsampling modules~\cite{Li2022ExploringPV}, and additional bias parameters~\cite{BitFit} to reduce training parameters during fine-tuning. 
Our work focuses on selecting the most semantically related part of the model and adapting to downstream tasks efficiently. No additional newly designed module is required.

\noindent \textbf{Continual learning.} 
Continual learning involves handling a diverse set of tasks and accumulating knowledge through a series of training.
Recent efforts have been made to address catastrophic forgetting, including imposing regularization~\cite{kirkpatrick2017overcoming,zenke2017continual,ritter2018online} and retaining a small buffer of data for replay~\cite{lopez2017gradient, v.2018variational}. Some approaches~\cite{yoon2017lifelong,hung2019compacting} dynamically expand the network by adding neurons to each MLP or convolution layer. In contrast, our modular design enables straightforward well-organized expansion by adding experts and learning new routers. Moreover, since each dataset has its router, the experts added will not be chosen by the previous dataset. Unlike other expansion techniques, our approach does not suffer from catastrophic forgetting.



\section{Method}

We start with the definition of multi-task heterogeneous training. Suppose we have $M$ datasets $D_1$, $D_2$, ..., $D_M$. Each dataset contains a set of training pair $\{I; T_i(I)\}$  and $T_i$ is the task on dataset $D_i$ that map images $I$ to $T_i(I)$. Here, we assume each dataset only has one task to do for simplicity. Multi-task heterogeneous training is to learn a joint model on the $M$ datasets at once.
 
\subsection{Preliminary}

\textbf{Mixture-of-Experts (MoE). }  A MoE layer contains a group of expert networks $E_1, E_2, ..., E_N$ and a routing network $G$. The routing network $G$ calculates the weight $G^k(x)$ for each expert $E_k$ given input $x$ and the output of an MoE layer is the weighted sum of the output of every expert $E_k(x)$.  Formally, the output of an MoE layer is 
\vspace{-0.1in}
\begin{align} \label{eqn:moe_output}
y =&\sum_{k=1}^{N} G^k(x)  E_{k}({x}).
\end{align}

The routing network $G$ is a Top-$K$ Routing network~\cite{shazeer2017} that only $K$ experts with the highest weight contribute to the final output:
\begin{align} \label{eqn:topk_output}
    G(x) =&  \operatorname{TopK}(\operatorname{Softmax} (xW_g), k)
\end{align}
where $\operatorname{TopK}(\cdot,k)$ sets all elements in the vector to zero except the elements with the largest $K$ values.

\textbf{Mutual information loss.} Mod-Squad~\cite{chen2022modsquad} proposes a mutual information loss as an auxiliary loss to better assign experts to {\em tasks} so that each expert is more likely to be used for a fixed set of tasks.
In contrast, the key motivation in \model is to encourage experts to specialize on {\em datasets} and then when adapting to downstream tasks, the downstream datasets are more likely to activate a small subset of experts. So we have $M$ \textit{dataset-specific routing networks} and modify the loss so that the experts are assigned to datasets instead of tasks:
\begin{align} \label{eqn:MI_split}
L_{MI} = -\sum_{i=1}^M \sum_{j=1}^K P(D_i,E_j) \log P(D_i,E_j) 
 + \sum_{i=1}^M P(D_i)  \log P(D_i) 
 + \sum_{j=1}^K P(E_j) \log P(E_j).
\end{align}
As in \cite{chen2022modsquad}, we assume that $P(D_i)=\frac{1}{M}$ as we want all datasets to be considered equally important. We have $P(E_j|D_i)=\sum_{x\in D_i}G_i^j(x)$
where $G_i^j$ is the weight of expert $E_j$ for dataset $D_i$. 
With $P(E_j|D_i)$, we can get $P(D_i,E_j)=P(E_j|D_i)P(D_i)$ and $P(E_j)=\sum_{i=1}^MP(D_i, E_j)$. 

\subsection{Multi-Task Heterogeneous Training}
 
\textbf{Backbone architecture.} Our multi-task heterogeneous training is a general framework that is orthogonal to model architecture design. All Transformer or MLP-based structures are applicable.
In this work, we choose two recent state-of-the-art transformer architectures for both image classification and dense prediction tasks as our backbone: Swin-Transformer~\cite{liu2021swin} and DaviT~\cite{ding2022davit}. We replace the MLP layers in these two models with MoE MLP layers.

\textbf{Task-specific module.}
Vision tasks require specific designs of modules to process the data and different ways of perception have a huge impact on performance.
While recent studies~\cite{lu2023unifiedio, jaegle2021perceiver} tend to use a shared task module for all tasks, we believe that the inherent differences in vision tasks make it difficult for a shared module to capture the essential information for all tasks. 
Thus, \model incorporates all task-specific designed modules (e.g., feature pyramid network), with only a backbone transformer shared among all tasks.

\textbf{Sampling strategy.} Data sampling plays a crucial role in heterogeneous training. 
Datasets could have varying scale levels with huge gaps in batch size.
For example, while a single GPU may work with 128 samples on image classification, it could only afford 2 samples on detection and segmentation.
Most multi-task frameworks~\cite{chen2022modsquad, xumtformer, kokkinos2017ubernet} tend to update the network after forwarding for all tasks.
However, such approaches are impractical as the GPU memory is heavily consumed when activating all dense vision modules, \eg, detection head and segmentation head. 
Also, forwarding samples from all tasks in one batch is not scalable when having more tasks.

To address the above issue, \model adopts a two-step sampling.
We first apply weighted sampling to select one out of the $M$ datasets, then randomly sample a batch of data from the chosen dataset.
The weight assigned to each dataset $D_i$ for sampling is denoted as $w_{sample_i}$, which can be pre-defined by the total number of iterations required for convergence in single dataset training, with some empirical tuning. 
Note that for relatively small datasets, a sufficiently large weight should be assigned to prevent degeneration caused by overtraining on other datasets. 
%

\textbf{Optimization and convergence.} 
Each task in our framework is associated with its unique module designed to process the data and its own loss.
The losses on datasets $D_i$ are weighted and alternately optimized with a predetermined weight $w_{l_i}$ for each dataset.
One challenge in optimization is the presence of gradient conflicts between different tasks. These conflicts
interfere with the joint optimization and slow down the convergence.
It is also not uncommon to observe that one task dominates the training process while others lag behind.
We find  that well-defined loss weights and sampling weights contribute to the stabilization of training, and the large batch optimizer Lamb~\cite{you2019large} works well in heterogeneous training.
Effective convergence in heterogeneous training requires approximately 50 percent more iterations than the combined number of iterations for each individual single-task training.
These additional training iterations account for the complexity introduced by joint optimization over diverse vision tasks.



\textbf{New mutual information loss for multi-task heterogeneous training.} 
In Mod-Squad~\cite{chen2022modsquad}, the mutual information loss in Equ.~\ref{eqn:MI_split} can be calculated in each batch as all tasks are contained in one batch. 
However, calculating $P(D,E)$ and $P(E)$ within a sampled batch from one random dataset in heterogeneous training leads to heavy bias.
To deal with this, we use an approximation inspired by the following idea:
\begin{align} \label{eqn:math_trick}
\frac{\partial}{\partial x}[x\log x] = 1 + \log x =  \frac{\partial}{\partial x}[(1+\log c)x]|_{c=x}.
\end{align}

This suggests that if we replace $x\log x$ with $(1+\log c)x$, and $c$ is a good approximation of $x$, then we will still have a similar gradient.
In our case, we will approximate a {\em running estimate} of the joint distribution of $P(D,E)$ with a buffer $B(D,E)$. The running estimate $B(D,E)$ avoids the heavy bias caused by estimating $P(D,E)$ from a single task data set. In each forward pass when we sample dataset $D_i$, we momentum update $B(D_i,E)$ with a momentum of $0.98$. This keeps the estimate of $B$ close to that of the desired joint distribution. 
Using this idea, we rewrite  Eq.~\ref{eqn:MI_split} and use the resulting equation as the loss function to calculate the gradient. The equation is given by:
\begin{align} \label{eqn:new}
L_{MI} = -\sum_{i=1}^M \sum_{j=1}^K [1+\log B(D_i,E_j)]P(D_i,E_j)
 + \sum_{j=1}^K [1+\log(\sum_{i=1}^MB(D_i,E_j))]P(E_j).  
\end{align}
Here, 
$P(D_i,E_j),P(E_j)$ is calculated in each forward pass backpropping gradients.  If $D_i$ is not sampled in the current forward pass, $P(D_i,E_j)$ is set to 0. 
Note that $P(D_i)\log P(D_i)$ is ignored as a constant. When adapting to new downstream datasets, the buffer still memorizes $P(D,E)$ for old datasets. Therefore, the MI loss can still be computed to balance experts on new datasets, which is not applicable in \cite{chen2022modsquad}.
\subsection{Efficient Adaptation on Downstream Tasks}

Mod-squad~\cite{chen2022modsquad} explores modular design in multi-task learning, which helps mitigate task conflicts and enables the extraction of sub-models for specific tasks.
However, learning from a single multi-label dataset, its applicability is limited to scenarios similar to the pre-trained dataset, thereby restricting its generalizability across diverse vision datasets.
Consequently, it's hard for the method to gain enough benefit from multi-task training, and its downstream performance is somehow restrained.
%

In comparison, we scale up multi-task learning to mainstream vision datasets, leading to better generalizations on downstream tasks.
Benefiting from the strong modularity, \model can be easily decomposed into high-performing components and also allows for a more flexible selection of semantically meaningful components when transferring to downstream tasks, ensuring efficient adaptation capabilities.

\model has two appealing advantages: \textbf{1) Downstream applications can select experts that best match the downstream scenario.}
This can be done by learning a new router in each MoE module to find good experts for the downstream task. We consider an expert as a good expert if it is chosen with a high frequency by the router on the downstream dataset. The routers are very lightweight (0.4M in parameters) and can quickly converge to the optimum while freezing all other parameters. 
\textbf{2) We can easily control the architecture within each small component (a small MoE module).} It is easy to expand or prune the model by simply adding or removing experts. 
This flexibility enables efficient customization of the model based on the specific requirements of the task at hand.


With these two advantages, we can do efficient fine-tuning in the following aspects as shown in Fig.~\ref{fig:adapt}: 
\textbf{1) fewer training parameters}. The model only needs to learn a new router for the downstream dataset and optionally fine-tune a few experts in each MoE module.
\textbf{2) fewer model parameters}. After learning a new router for the new downstream dataset, we can rank experts according to the frequency chosen by the routers. Then, we can remove some of the experts rarely being used.
\textbf{3) lower computation cost}. The new router for the downstream dataset can be learned with a smaller Top-K. So that fewer experts are chosen during one forward pass and can greatly reduce the computation cost and inference latency.
Note that all these ways of efficient adaptation can be combined together to meet the demands of downstream datasets.



\subsection{Continual Learning}

The strong modularity also enables simple model expansion and helps conduct continual learning. Specifically, we directly add $C$ experts in each MoE module along with new task-specific routers every time learning a new task. We train on the new task but freeze all parameters except for the newly added part.
There are three main advantages of this approach: 1) No catastrophic forgetting. As all the experts are unchanged after learning and the newly added experts will not be chosen by the router of previous tasks, there is no catastrophic forgetting. 2) Well-organized architecture and knowledge reuse. The model still keeps an elegant modularized design. The routers select experts to reuse knowledge related to the new task and ignore experts with unrelated expertise.
3) The computation cost is constant. Other expanding methods~\cite{yoon2017lifelong,hung2019compacting} add both computation cost and capacity to the existing model, while our approach only adds capacity. This makes our approach expandable with a large number of tasks.


\section{Experiments}
\subsection{Multi-task heterogeneous training.}
We conduct three fundamental vision tasks (classification, detection, and segmentation) on three datasets: ImageNet-1K~\cite{deng2009imagenet}, COCO~\cite{COCO}, and ADE20K~\cite{zhou2017scene}. For the downstream datasets, we evaluate classification on the scene dataset Places-365~\cite{zhou2017places} (P365), the popular fine-grained dataset iNaturalist-2018~\cite{van2018inaturalist} (iNat18), the pet dataset Pets~\cite{parkhi2012cats}, the fine-grained bird dataset CUB~\cite{wah2011caltech}, and the car dataset Cars~\cite{krause20133d}. We evaluate downstream detection on PASCAL VOC~\cite{everingham2010pascal} and downstream segmentation on Cityscapes~\cite{cordts2016cityscapes} and NYU~\cite{silberman2012indoor}.


\textbf{Models and baselines.} 
We utilize Swin Transformer~\cite{liu2021swin} and DaViT~\cite{ding2022davit} as our backbone transformers, with results reported on three different sizes: tiny (T), small (S), and base (B).
Each task has its own task-specific head. For classification, we use a single linear layer. For detection, we use the retina head~\cite{lin2017focal}. For segmentation, we use the UperNet~\cite{xiao2018unified}. Each task follows its own input and output format based on single-task methods.
We implement our methods and baselines as the following:
1)  Train from scratch (Scratch): a vanilla single-task learning baseline that trains models from scratch.
2) Pre-train then fine-tune (Pre. \& FT.): pre-training on ImageNet followed by fine-tuning on the target dataset.
3) \model.D: our multi-task heterogeneous learner using a dense model (no MoE).
4) \model: our multi-task heterogeneous learner using a sparse model (with MoE).


\textbf{Configs.} We employ 12 experts with Top-K as 4 for all MoE modules, following~\cite{chen2022modsquad}. For base-size transformers, we replace the MLP with MoE MLP every 2 transformer layers. For small and tiny transformers, we use MoE Mlp in every transformer layer.
All models are trained for 240,000 iterations on 96 Tesla V100 GPUs with Lamb~\cite{you2019large} as the optimizer for large batch training. Weight decay is set to 0.05 and the maximal gradient norm is clipped to 0.1. We use a simple triangular learning rate schedule with a maximum learning rate of 0.004, as in~\cite{ding2022davit}. Data augmentations for each task follow the common practice in~\cite{liu2021swin,ding2022davit}.
During multi-task heterogeneous training, data sampling weight is set to \{3, 2, 1\}, loss weight is set to \{1.0, 0.6, 0.2\}, and batch size is set to \{64, 2, 2\} for classification, detection, and segmentation, respectively. For a fair comparison, all results of our method and baselines are obtained from our implementations with the same settings.
More details of the training settings, models, and datasets can be found in the Appendix.

\begin{table*}[t]
\begin{center}
\footnotesize
\tabcolsep=0.1cm
\caption{\textbf{Multi-task heterogeneous training.} We compare it with training from scratch (scratch) and pre-training then fine-tuning (pre. \& ft.). Note that on COCO and ADE20K, pre. \& ft. would initialize the backbone with an IN-1K pre-trained model. 
The numbers of parameters and FLOPs of the backbone are measured. Note that all classifications have input resolution $224\times 224$.
The single-crop test is used for semantic segmentation.
}
\label{tab:mtl}
\vspace{-0.1in}
\begin{tabular}{c|c|cc|cc|ccc|ccc}
\toprule
\multirow{2}{*}{Backbone} & \multirow{2}{*}{Model} & \multirow{1}{*}{Params} & \multirow{1}{*}{FLOPs} &\multicolumn{2}{c|}{\cls{\textbf{IN-1K}}} & \multicolumn{3}{c|}{\dete{\textbf{COCO}}} & \multicolumn{3}{c}{\seg{\textbf{ADE20K}}} \\ 
& & (M) & (G) & \multirow{1}{*}{\cls{\textbf{top-1}}} & \multirow{1}{*}{\cls{top-5}}  & \multirow{1}{*}{\dete{$\mathbf{mAP}$}} & \multirow{1}{*}{\dete{$mAP_{50}$}} & \multirow{1}{*}{\dete{$mAP_{75}$}} & \multirow{1}{*}{\seg{$\mathbf{mIoU}$}} & \multirow{1}{*}{\seg{$mAcc$}} & \multirow{1}{*}{\seg{$aAcc$}} \\
\midrule      
\multirow{3}{*}{Swin-T} & Scratch & 27.5$\times$3 & 4.4 & \textbf{80.6} & 95.2 & 34.9 & 54.3 & 36.6 & 32.0 & 41.4 & 75.8 \\
                       &  Pre. \& FT. & 27.5$\times$3 & 4.4 & -- & -- & 42.0 & 64.7 & 45.9 & 44.3 & 55.8 & 81.0\\
                       & \model.D & 27.5 & 4.4 & 79.7 & 95.1 & 43.8 & 65.7 & 46.8 & 44.4 & 54.8 & 80.5 \\
                       & \model & 50.9 & 5.1 & 80.3 & 94.7 & \textbf{45.0} & 66.5 & 48.2 & \textbf{44.6} & 55.0 & 81.0 \\
\hline
\multirow{3}{*}{Swin-S} & Scratch & 48.9$\times$3 & 8.5 & \textbf{82.6} & 96.1 & 36.3 & 55.6 & 38.4 & 34.5 & 43.9 & 77.1\\
                       &  Pre. \& FT. & 48.9$\times$3 & 8.5 & -- & -- & \textbf{46.0} & 68.0 & 49.9 & 47.0 & 56.9 & 81.7 \\
                       & \model.D & 48.9 & 8.5 & 80.7 & 95.5 & 45.8 & 67.8 & 48.7 & \textbf{47.7} & 58.4 & 81.8 \\
                       & \model & 89.1 & 9.2 & 82.0 &  95.9 & 45.7 & 66.8 & 49.1 & 46.7 & 57.1 & 81.8 \\
\hline
\multirow{3}{*}{Swin-B} & Scratch & 86.7$\times$3 & 15.1 & \textbf{83.1} & 96.4 & 35.5 & 54.7 &  37.4 & 35.4 & 44.8 & 77.6 \\
                       &  Pre. \& FT. & 86.7$\times$3 & 15.1 & -- & -- & 47.3 & 69.0 & 51.2 & 47.7 & 58.7 & 82.3 \\
                       & \model.D & 86.7 & 15.1 & 82.2 & 96.2 & 47.5 & 69.2 & 51.0 & \textbf{48.8} & 59.7 & 82.5 \\
                       & \model & 158.3 & 16.2 & 82.3 & 96.2 & \textbf{47.6} & 69.1 & 50.9 & 48.2 & 59.0 & 82.5 \\
\hline
\hline
\multirow{3}{*}{DaViT-T} & Scratch & 27.6$\times$3 & 4.4 & \textbf{82.5} & 96.2 & 37.7 & 57.1 & 40.0 & 36.4 & 46.4 & 77.8 \\
                       & Pre. \& FT. & 27.6$\times$3 & 4.4 & -- & -- & \textbf{45.4} & 66.9 & 48.4 & 45.8 & 56.0 & 81.8 \\
                       & \model.D & 27.6 & 4.4 & 81.3 & 95.8 & 44.6 & 66.6 & 47.5 & 46.2 & 56.4 & 81.6 \\
                       & \model & 51.2 & 5.1 & 82.0 & 95.8 & 45.1 & 67.5 & 48.1 & \textbf{47.4} & 57.1 & 82.1 \\
\hline
\multirow{3}{*}{DaViT-S} & Scratch & 49.0 $\times$3 & 8.6 & \textbf{83.8} & 96.8 & 37.8 & 56.7 & 40.5 & 38.2 & 48.4 & 78.8 \\
                       & Pre. \& FT. & 49.0 $\times$3 & 8.6 & -- & -- &  47.2 & 68.9 & 50.7 & 48.3 & 60.2 & 82.3 \\
                       & \model.D &  49.0 & 8.6 & 82.6 & 96.5 & \textbf{47.3} & 69.2 & 50.6 & \textbf{48.7} & 59.1 & 82.7 \\
                       & \model & 88.9 & 9.2 & 83.3 & 96.5 & 46.4 & 67.7 & 49.5  & 47.6 & 57.9 & 82.6 \\
\hline
\multirow{3}{*}{DaViT-B} & Scratch & 86.9$\times$3 & 15.2 & \textbf{84.2} & 96.9 & 38.0 & 57.2 & 40.5 & 38.5 & 48.7 & 78.9 \\
                       & Pre. \& FT. & 86.9$\times$3 & 15.2 & -- & -- & 48.1 & 69.7 & 51.3 & 49.3 & 60.2 & 83.0 \\
                       & \model.D & 86.8 & 15.2 & 83.9 &  96.9 & \textbf{48.3} & 70.4 & 51.8 & \textbf{50.0} & 60.3 & 83.1 \\
                       & \model & 158.7 & 16.3 & 83.6 & 96.8 &  47.8 & 69.5 & 51.5 & 49.6 & 60.1 & 83.1 \\
\bottomrule
\end{tabular}
\end{center}
\vspace{-12pt}
\end{table*}



\begin{table*}[t]
\begin{center}
\footnotesize
\tabcolsep=0.163cm
\caption{\textbf{Comparisons of different pre-training schemes on downstream performance.} We compare with IN-1K pre-trained model (IN-1K Pre.), and multi-task multi-label pre-training (Mod-Squad~\cite{chen2022modsquad}) on Taskonomy~\cite{zamir2018taskonomy}. To calculate the mean, we first average the performance on classification, detection, and segmentation separately. Afterward, we average the results across all tasks.
} 
\label{tab:pretrain}
\vspace{-0.1in}
\begin{tabular}{c|c|ccccc|c|cc|c}
\toprule
\multirow{2}{*}{Backbone} & \multirow{2}{*}{Method} & \multicolumn{1}{c}{\cls{P365}} & \multicolumn{1}{c}{\cls{iNat18}} & \multicolumn{1}{c}{\cls{Pets}} & \multicolumn{1}{c}{\cls{CUB}} & {\cls{Cars}}  & {\dete{PASC.}}  & \multicolumn{1}{c}{\seg{City.}} & {\seg{NYU}} & \multirow{2}{*}{\textbf{Mean}}\\ 

& & {\cls{top-1}} & {\cls{top-1}} & {\cls{top-1}} & {\cls{top-1}} & {\cls{top-1}} &  {\dete{$mAP$}}  & {\seg{$mIoU$}} & {\seg{$mIoU$}}
\\
\midrule 
\multirow{4}{*}{Swin-B} 
                       & IN-1K Pre. 
                        & 58.7 & 72.9 & 94.0 & 83.9 & 94.0 & 76.9 & 80.6 & 76.2 & \cellcolor{mygray}78.7
                       \\
                       & Mod-Squad~\cite{chen2022modsquad} & 56.4 & 69.4 & 92.3 & 79.8 & 93.7 & 77.2 & 81.1 & 77.5 & \cellcolor{mygray}78.1\\
                       & \model.D & 
                       59.1 & 73.3 & 94.2 & 84.3 & 94.2 & 78.7 & 82.1 & 78.0 & \cellcolor{mygray}79.9
                       \\ 
                       & \model & 59.4 & 73.6 & 94.6 & 84.7 & 94.9 & 79.1 & 82.5 & 78.7 & \cellcolor{mygray}\textbf{80.4}\\
\midrule
\multirow{3}{*}{Davit-B} 
                       & IN-1K pre. 
                        & 59.2 & 73.4 & 94.4 & 88.4 & 94.9 & 77.4 & 81.5 & 76.7 & \cellcolor{mygray}79.5
                       \\
                       & \model.D 
                       & 59.6 & 73.5 & 94.8 & 89.0 & 95.0 & 78.8 & 82.7 & 78.6 & \cellcolor{mygray}80.6 
                       \\
                       & \model 
                       & 60.1 & 73.9 & 94.9 & 89.4 & 95.0 & 79.5 & 83.4 & 79.3 & \cellcolor{mygray}\textbf{81.2}
                       \\ 
\bottomrule
\end{tabular}
\end{center}
\vspace{-18pt}
\end{table*}

\textbf{Multi-task heterogeneous training.} 
We compare different training schemes as shown in Tab.~\ref{tab:mtl}. 
Across all three datasets with varying backbones, we observe that:
1) Heterogeneous training performs on par with the state-of-the-art pre-train then fine-tune learning scheme, indicating the gradient conflicts between different tasks are alleviated by our modular design.
2) Notably, for the segmentation task, \model consistently outperforms the previous state-of-the-art across all backbone choices, suggesting that joint training with classification and detection tasks improves segmentation. 
3) \model also works pretty well on image detection and is superior to previous arts in most cases.
4) The \model and  \model.D generally exhibit similar performance on small and base models and  \model consistently outperforms \model.D on tiny models, likely influenced by the relationship between model capacity and dataset scale.



\textbf{Downstream performance.} As shown in Tab.~\ref{tab:pretrain}, we compare different training schemes on the downstream datasets. 
\model outperforms the single-task pre-trained model IN-1K Pre. and multi-task multi-label pre-trained model Mod-Squad, particularly on detection and segmentation tasks.
We also note that the sparse model \model consistently outperforms the dense model \model.D, indicating that additional experts for selection could be beneficial for downstream tasks.


\begin{table*}[t]
\begin{center}
\footnotesize
\tabcolsep=0.05cm
\caption{\textbf{Efficient adaptation.} All experiments use \model as the pre-trained model with Davit-S as the backbone. The ratio calculates the percentage of efficiency metric compared to the fully fine-tuned baseline. 
Notations: `Ro.' for Router, `Ex.' for expert(s), $\theta$ is a threshold on the frequency used for an expert.
We have two hybrid models: 1) `Hybrid-A' directly combines `Ro. w/ 1 Ex.', `Prune 2/3 Ex.', and `Top-K=2'. 2) `Hybrid-B' combines `Ro. w/ 2 Ex.', `Prune 2/3 Ex.', and `Top-K=3'. }
\label{tab:adaption}
\vspace{-0.1in}
\begin{tabular}{l|cccc|ccccc|c|cc|c}
\toprule
\multirow{2}{*}{Method} & \multirow{1}{*}{Train.} & \multirow{1}{*}{Model} & FLOPs & \multirow{2}{*}{Ratio} & \multicolumn{1}{c}{\cls{P365}} & \multicolumn{1}{c}{\cls{iNat18}} & \multicolumn{1}{c}{\cls{Pets}} & \multicolumn{1}{c}{\cls{CUB}} & {\cls{Cars}}  & \multicolumn{1}{c|}{\dete{PASC.}}  & \multicolumn{1}{c}{\seg{City.}} & {\seg{NYU}} 
& \multirow{2}{*}{\textbf{Mean}}
\\ 
& Par.(M) & Par.(M) & (G) &  & \multicolumn{1}{c}{\cls{top-1}} & \multicolumn{1}{c}{\cls{top-1}} & \multicolumn{1}{c}{\cls{top-1}} & \multicolumn{1}{c}{\cls{top-1}} & {\cls{top-1}} &  \multicolumn{1}{c|}{\dete{$mAP$}}  & \multicolumn{1}{c}{\seg{$mIoU$}} & {\seg{$mIoU$}}
\\
\midrule      
                        FT-Full & 88.9 & 88.9 & 9.2 & - & 59.0 & 72.9 & 94.0 & 88.2 & 95.0 & 78.6 & 81.4 & 77.4 & \cellcolor{mygray}79.9\\
\midrule
                       Adapter~\cite{houlsby2019parameter} & \cellcolor{mygray}14.8 & - & - & 16.6\% & 50.7 & 62.4 & 81.1 & 75.8 & 80.8 & 67.7 & 69.9 & 66.8 & \cellcolor{mygray}68.7\\ 
                       Ro. Only & \cellcolor{mygray}\textbf{0.4} & - & - & 0.4\% & 52.1 & 64.2 & 83.3 & 77.9 & 78.2  & 69.6 & 71.8 & 68.7 & \cellcolor{mygray}70.3\\
                       Ro. w/ 1 Ex. & \cellcolor{mygray}5.4 & - & - & 6.1\% & 57.4 & 70.7 & 91.3 & 85.8 & 94.7 & 76.5 & 78.8 & 75.2 & \cellcolor{mygray}77.8\\
                       Ro. w/ 2 Ex. & \cellcolor{mygray}10.4 & - & - & 11.7\% & 58.8 & 72.7 & 94.0 & 87.8 & 95.0 & 77.9 & 80.7 & 76.7 &  \cellcolor{mygray}\textbf{79.4}\\
\midrule            
                    Prune $\theta=1\%$ & - & \cellcolor{mygray}60.2 & - & 67.7\% & 58.9 & 72.8 & 93.9 & 88.1 & 95.0 & 78.6 & 81.4 & 77.3 & \cellcolor{mygray}\textbf{79.9}\\
                    Prune $\theta=5\%$ & - & \cellcolor{mygray}54.4 & - & 61.2\% & 58.8 & 72.7 & 93.8 & 88.0  & 94.  & 78.4 & 81.4 & 77.2 & \cellcolor{mygray}79.7\\
                    Prune 1/2 Ex. & - & \cellcolor{mygray}59.9 & - & 67.3\% & 58.8 & 72.8 & 93.9 & 88.0  & 93.9 & 78.6 & 81.4 & 77.3 & \cellcolor{mygray}79.8\\
                    Prune 2/3 Ex. & - & \cellcolor{mygray}\textbf{49.9} & - & 56.1\% & 58.8 & 72.6 & 93.6 & 87.8 & 93.8 & 78.6 & 81.3 & 77.2 & \cellcolor{mygray}79.7\\
\midrule
                    Top-K=3 & - & - & \cellcolor{mygray}7.7 & 83.7\% & 58.8 & 72.5 & 93.3 & 87.3 & 94.9 & 77.3 & 80.1 & 76.3 & \cellcolor{mygray}\textbf{79.0}\\
                    Top-K=2 & - & - & \cellcolor{mygray}6.2 & 67.4\%\ & 58.1 & 70.7 & 91.9 & 86.2 & 92.0 & 74.9 & 77.6 & 73.7 & \cellcolor{mygray}76.8\\
                    Top-K=1 & - & - & \cellcolor{mygray}\textbf{4.7} & 51.0\%\ & 48.5 & 59.9 & 77.3 & 72.4 & 77.4 & 64.3 & 66.6 & 63.3 & \cellcolor{mygray}65.4\\

\midrule

Hybrid-A & \cellcolor{mygray}5.4 & \cellcolor{mygray}49.9 & \cellcolor{mygray}6.2 & - & 58.0 & 70.6 & 91.1 & 85.8 & 94.7 & 76.3 & 78.5 & 73.2 & \cellcolor{mygray}77.4
\\
Hybrid-B & \cellcolor{mygray}10.4 & \cellcolor{mygray}49.9 & \cellcolor{mygray}7.7 & - & 58.8 & 72.4 & 93.3 & 87.2 & 94.9 & 77.1 & 79.9 & 76.2 & \cellcolor{mygray}\textbf{78.8}
\\
\bottomrule
\end{tabular}
\end{center}
\vspace{-20pt}
\end{table*}

\subsection{Efficient Adapters}

In this section, we highlight the potential of \model as an efficient adapter.

\textbf{Efficient in training parameters.} 
\model can adapt quickly to a new task or dataset by tuning the router with a few optional experts and learning a new task head. During this process, all other parameters are frozen. The optional few experts to be fine-tuned are randomly selected. We find that randomly selected experts perform similarly to selecting the expert with the highest or lowest use frequency on the downstream dataset. Please refer to the supplementary for more details.

\begin{figure}[t]
  \centering
  \includegraphics[width=0.99\textwidth]{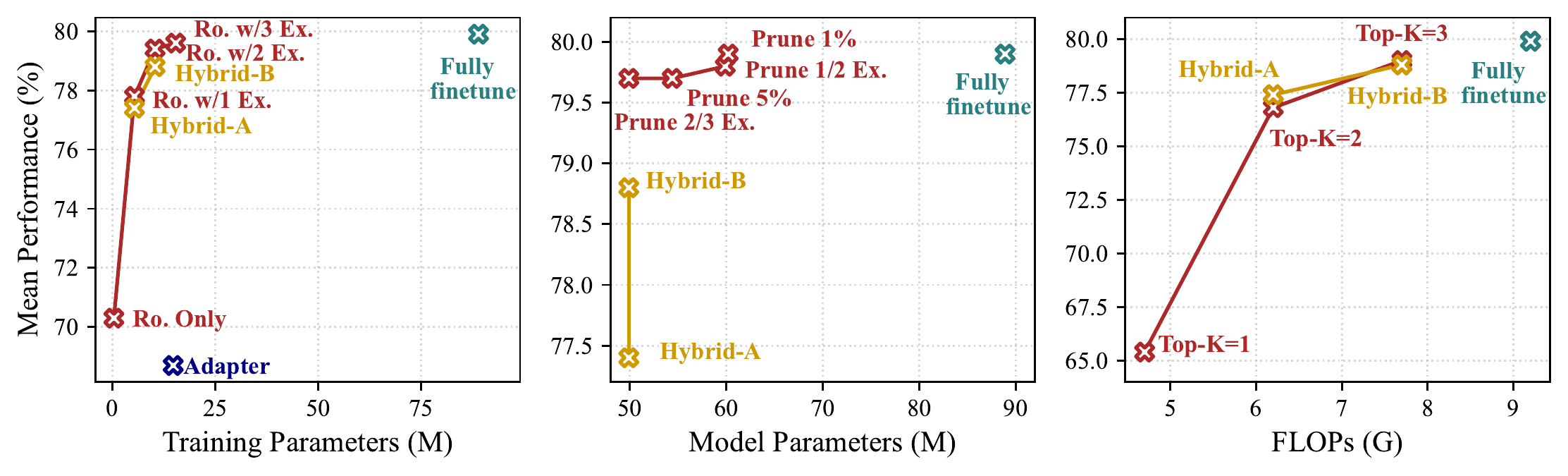}
  \vspace{-0.1in}
  \caption{ \textbf{Trade-off between efficiency and performance.} We visualize the trade-off between performance and training parameters, model parameters, and computation cost respectively. 
  }
  \vspace{-10pt}
  \label{fig:comparison}
\end{figure}

In Tab.~\ref{tab:adaption}, our method is referred to as 'Ro. Only', 'Ro. w/ 1 Ex.', and 'Ro. w/ 2 Ex.' that denotes only tuning routers, routers with one expert per MoE module, and routers with  two experts per MoE module, respectively.
We compare our efficiency in training parameters with the commonly used adapter~\cite{houlsby2019parameter}, which adds an adapter module after each MoE MLP block. In contrast, we only need new lightweight routers (0.4M) and one or two additional experts per MoE module. Even updating only new routers outperforms the adapter baseline, and Ro. w/2 Ex. has a very close performance (0.5 points lower in mean) to the fully fine-tuned baseline. For a clearer comparison, please see Fig.~\ref{fig:comparison}.

\textbf{Efficient in model capacity.}  
In terms of model capacity, \model can remove experts after learning a new router on the new task. This can be achieved by removing experts with the least use frequency, followed by fine-tuning the entire model.

We explore two methods of pruning: 1) Removing a few experts from each MoE layer. In Tab.~\ref{tab:adaption}, we attempt to remove 1/2 experts and 2/3 experts.
2) Removing all experts whose use frequency is lower than a threshold $\theta$ on the downstream dataset. This approach may result in a different number of experts in each MoE layer, but it has comparable efficiency to the first pruning method.
Results and a clear comparison can be referred to Tab.~\ref{tab:adaption} and Fig.~\ref{fig:comparison}.

\begin{table*}[t]

\begin{center}
\caption{\textbf{Continual learning.} We conduct continual learning on these datasets one by one after heterogenous pre-training and report the final performance.
All experiments use \model as the pre-trained model with DaviT-S as the backbone. The number of training parameters and newly added parameters in the backbone per task are measured. Here the average is the average performance on all datasets.
} 
\vspace{-0.1in}
\label{tab:continual_learning}
\footnotesize
\tabcolsep=0.07cm
\begin{tabular}{c|cc|cccccccc|c}
\toprule
\multirow{2}{*}{Method} & \multirow{1}{*}{New params}  & \multirow{1}{*}{Train. params} & \multicolumn{1}{c}{\cls{P365}} & \multicolumn{1}{c}{\cls{iNat18}} & \multicolumn{1}{c}{\cls{Pets}} & \multicolumn{1}{c}{\cls{CUB}} & {\cls{Cars}}  & \multicolumn{1}{c}{\dete{PASC.}}  & \multicolumn{1}{c}{\seg{City.}} & {\seg{NYU}} 
& \multirow{2}{*}{\textbf{Average}}

\\ 
& per task (M) & per task (M) & \multicolumn{1}{c}{\cls{top-1}} & \multicolumn{1}{c}{\cls{top-1}} & \multicolumn{1}{c}{\cls{top-1}} & \multicolumn{1}{c}{\cls{top-1}} & {\cls{top-1}} &  \multicolumn{1}{c}{\dete{$mAP$}}  & \multicolumn{1}{c}{\seg{$mIoU$}} & {\seg{$mIoU$}}

\\
\midrule      
LWF~\cite{kirkpatrick2017overcoming} & 0 & \cellcolor{mygray}88.9 & 46.2 & 57.0  & 73.5 & 70.6 & 75.5 & 62.7 & 71.1 & 68.9 & \cellcolor{mygray}65.7
\\
Rou. only & 0.4 & \cellcolor{mygray}\textbf{0.4} & 52.1 & 64.2 & 83.3 & 77.9 & 78.2  & 69.6 & 71.8 & 68.7 & \cellcolor{mygray}70.7
\\
Rou. w/ 1Ex. & 5.4 & \cellcolor{mygray}5.4 & 57.6 & 70.8 & 91.3 & 85.9 & 94.7 & 76.8 & 79.0 & 75.6 & \cellcolor{mygray}79.0
\\
Rou. w/ 2Ex. & 10.4 & \cellcolor{mygray}10.4 & 58.8 & 72.8 & 94.5 & 88.0 & 95.0 & 78.1 & 80.7 & 76.9 &  \cellcolor{mygray}\textbf{80.6}
\\
\midrule 
FT-Full & -- & -- & 59.0 & 72.9 & 94.0 & 88.2 & 95.0 & 78.6 & 81.4 & 77.4 & \cellcolor{mygray}80.8\\

\bottomrule
\end{tabular}
\end{center}
\vspace{-20pt}
\end{table*}

\textbf{Efficient in computation cost.} 
Most pre-training may use a relatively large backbone, but the downstream tasks/datasets may not require such a large model capacity.
MTHT.S can regulate the computation cost by learning new routers with a reduced Top-K. This would result in a trade-off between performance and computation cost, as illustrated in Fig.~\ref{fig:comparison}. For some datasets (\eg, P365), it can achieve a relatively low computation cost (\eg, 67.4\%) while maintaining the same level of performance (\eg, <1\% drop).

\textbf{Combine all efficient adapting.} 
To further improve efficiency, the efficient adapting techniques mentioned above can be combined. In Tab.~\ref{tab:adaption}, for Hybrid-B, we first learn a new router and remove 2/3 experts. Then, we fine-tune the router with Top-K as 3 along with two experts per module. This approach achieves a mean performance of 78.8, which is only 1 point lower than fine-tuning the entire model. Moreover, this method reduces training parameters, model parameters, and computation cost simultaneously.





\subsection{Continual learning.}
Continual learning without any forgetting is achievable with \model by learning new routers (0.4M) and a few optional experts on the new dataset. We compared it with the common regularization-based continual learning baseline LWF\cite{kirkpatrick2017overcoming}. As demonstrated in Tab.~\ref{tab:continual_learning}, our method has three significant advantages: 1) No forgetting on the learned datasets. 2) Only a smart part of the model needs to be trained on new datasets, requiring only 10.4M training parameters, while LWF needs to tune the whole model (88.9M). 3) Comparable performance to fully fine-tuning the whole model on every dataset.


\section{Conclusion}

Our study focuses on multi-task heterogeneous training and its adaptation ability on downstream datasets. MTHL can achieve outcomes comparable to the previous single-task state-of-the-art on all tasks. Furthermore, we investigate various methods of utilizing modularity to efficiently adapt to downstream tasks. Modularity also allows model expansion easily for continual learning. 
The broader impact of our work could be significant in terms of advancing general-purpose vision model pre-training and effective adaptation of large-scale models. One limitation of MTHL is model may be biased toward certain datasets and require more training iterations for convergence.






{
\bibliographystyle{abbrv}
\bibliography{ref2}
}

\newpage
\appendix

\setcounter{table}{0}
\setcounter{figure}{0}
\renewcommand{\thetable}{A\arabic{table}}
\renewcommand{\thefigure}{A\arabic{figure}}
\renewcommand{\thesubsection}{A\arabic{subsection}}
\section*{Appendix}

\renewcommand\twocolumn[1][]{#1}

\maketitle

\begin{center}
    \captionof{table}{
    We compare three ways of selecting a subset of experts to fine-tune, while freezing the remaining experts. We first learn new routers on the new downstream to determine each expert's frequency of being chosen.
    Random represents randomly choosing experts. Best represents choosing the experts with the highest frequency. Worse represents choosing the experts with the lowest frequency. We report mean top-1 accuracy on CUB, Cars, and Pets. Other settings are the same as in Tab.~\ref{tab:adaption} in the paper.
    }
    \centering
    {
    \begin{tabular}[h]{c|c|c|c}
        
          &   Random & Best & Worse  \\ \hline
        Ro. w/1 Ex.        &  \cellcolor{mygray}90.6   &   90.5 & 90.6          \\
        Ro. w/2 Ex.        &  \cellcolor{mygray}92.3  &   92.3 &  92.2             \\
    \end{tabular}
    }
    \label{tab:select}

\end{center}

\subsection{Different ways to select experts to be fine-tuned.} 
Tab.~\ref{tab:select} compares various methods of selecting experts to fine-tune while freezing the rest.
We compare random selecting experts and selecting experts that are more or less likely to be chosen by routers.
We find out that the selection method does not significantly affect the fine-tuning performance. Therefore, we use random selection for simplicity.

\subsection{Ablation on Top-$K$.}
\label{sec:expert}
As shown in Tab.~\ref{tab:topk}, we explore the effect on Top-$K$ in MoE module. 
The experiment setting is the same as in Table.1 in the paper with 12 experts per MoE module. 
We report the mean performance on pre-train and downstream datasets of our MHTL with Davit-T as the backbone.
To control the FLOPs to be the same for different Top-$K$, the hidden dimension of MLP experts is divided by $K$. 
All experiments have the same parameter size and the same FLOPs. We find that Top-$K=4$ has the best performance.

\subsection{Ablation on the number of experts.}

As shown in Tab.~\ref{tab:expert}, we explore the effect on the number of experts $E$ for the MoE MLP layer. 
The settings are the same as in \cref{sec:expert} with a Top-$K$ as 4.

\begin{table*}[h]
    \centering
    \small
    \caption{\textbf{Ablation study of Top-$K$ on MoE MLP layer.}}
    \label{tab:topk}
    \setlength{\tabcolsep}{0.9mm}{
        \centering
        \small    
        \begin{tabular}[h]{c|c|c|c|c|c}
        
          &    FLOPs(G)     & Params(M) & Hidden Dim &  Pre-train mean & Downstream mean  \\ \hline
        {K=2}        &   5.1   &    51.2 & 768 &  58.1 & 80.3          \\
        \rowcolor{mygray}
        {K=4}        &   5.1   &    51.2 & 384 & 58.2 & 80.4            \\
        {K=6}        &   5.1   &    51.2 &  256 &   57.9 & 80.0            \\
        \end{tabular}
    }

\end{table*}

\begin{table*}[h]
    \centering
    \small
    \caption{\textbf{Ablation study of expert number $E$ on MoE MLP layer.}}
    \label{tab:expert}
    \setlength{\tabcolsep}{0.9mm}{
        \centering
        \small    
        \begin{tabular}[h]{c|c|c|c|c|c}
        
          &    FLOPs(G)     & Params(M) &  Pre-train mean & Downstream mean  \\ \hline
        {E=6}        &   5.1   &    33.4 &  57.2 & 78.5          \\
        {E=9}        &   5.1   &    42.3 & 57.9 & 80.0            \\
        \rowcolor{mygray}
        {E=12}        &   5.1   &    51.2 &   58.2 & 80.4            \\
        {E=15}        &   5.1   &    60.1 &   58.2 & 80.5            \\
        \end{tabular}
    }

\end{table*}

\end{document}